\documentclass[12pt, final]{l4dc2022}
\usepackage{mathrsfs}
\usepackage{mathtools}
\usepackage{nicematrix}
\NiceMatrixOptions{renew-dots, renew-matrix, nullify-dots}

\newcommand{\bn}{\ensuremath{\boldsymbol{n}}}
\newcommand{\bv}{\ensuremath{\boldsymbol{v}}}
\newcommand{\bF}{\ensuremath{\boldsymbol{F}}}
\newcommand{\bJ}{\ensuremath{\boldsymbol{J}}}
\newcommand{\bV}{\ensuremath{\boldsymbol{V}}}
\newcommand{\bw}{\ensuremath{\boldsymbol{w}}}
\newcommand{\bx}{\ensuremath{\boldsymbol{x}}}
\newcommand{\bphi}{\ensuremath{\boldsymbol{\phi}}}
\newcommand{\bpsi}{\ensuremath{\boldsymbol{\psi}}}
\newcommand{\blambda}{\ensuremath{\boldsymbol{\lambda}}}
\newcommand{\btheta}{\ensuremath{\boldsymbol{\theta}}}
\newcommand{\bPhi}{\ensuremath{\boldsymbol{\Phi}}}
\newcommand{\bLambda}{\ensuremath{\boldsymbol{\Lambda}}}

\title[Adjoint Learning for TDKS]{Dynamic Learning of Correlation Potentials for a Time-Dependent Kohn-Sham System}
\usepackage{times}
\author{%
 \Name{Harish S. Bhat} \Email{hbhat@ucmerced.edu}\\
 \addr Department of Applied Mathematics, University of California Merced
 \AND
 \Name{Kevin Collins} \Email{kcollins9@ucmerced.edu}\\
 \addr Department of Physics, University of California Merced
 \AND
 \Name{Prachi Gupta} \Email{pgupta11@ucmerced.edu}\\
  \addr Department of Applied Mathematics, 
   \addr Department of Chemistry and Biochemistry, University of California Merced
    \AND
 \Name{Christine M. Isborn} \Email{cisborn@ucmerced.edu}\\
 \addr Department of Chemistry and Biochemistry, University of California Merced
}

\begin{document}

\maketitle
\vspace{-0.5cm}

\begin{abstract}
We develop methods to learn the correlation potential for a time-dependent Kohn-Sham (TDKS) system in one spatial dimension.  We start from a low-dimensional two-electron system for which we can numerically solve the time-dependent Schr\"odinger equation; this yields electron densities suitable for training models of the correlation potential.  We frame the learning problem as one of optimizing a least-squares objective subject to the constraint that the dynamics obey the TDKS equation.  Applying adjoints, we develop efficient methods to compute gradients and thereby learn models of the correlation potential.  Our results show that it is possible to learn values of the correlation potential such that the resulting electron densities match ground truth densities.  We also show how to learn correlation potential functionals with memory, demonstrating one such model that yields reasonable results for trajectories outside the training set.
\end{abstract}



\section{Introduction}

The time-dependent Schr{\"o}dinger equation (TDSE) governs the behavior of $N$ quantum particles, 
\begin{equation}
\label{eqn:tdse}
i \partial_t \Psi(r_1, r_2, ..., r_N,t) = \hat{H}(r_1, r_2, ..., r_N,t) \Psi(r_1, r_2, ..., r_N,t),
\end{equation}
where $\hat{H}$ is the Hamiltonian and $\Psi$ is the many-body wave function. In $d$-dimensional space, the many-body Coulomb interaction in the potential term of $\hat{H}$ leads to a coupled system of partial differential equations (PDE) in $dN+1$ variables.  Hence (\ref{eqn:tdse}) can only be solved for simple model problems, such as for one electron in three dimensions or two electrons in one dimension.  To simulate electron dynamics in molecules and materials, a widely used approach is time-dependent density functional theory (TDDFT), in which the many-body wave function $\Psi$ is replaced with the Kohn-Sham wave function $\Phi(r)$ to give the time-dependent Kohn-Sham (TDKS) equation \citep{maitra2016perspective, ullrich2011time-dependent}:
\begin{equation}
\label{eqn:tdks3d}
i \partial_t \Phi (r,t) = \sum_{i=1}^N [ -(1/2)\nabla^2_i  + v^{\text{ext}}(r_i,t)  + v^H[n](r_i,t)  + v^{XC}[n, \Psi_0, \Phi_0](r_i,t) ] \Phi(r,t).
\end{equation}
Because $\Phi(r)$ is constructed as a product of non-interacting single-particle orbitals $\phi(r_i)$, (\ref{eqn:tdks3d}) decouples into $N$ separate evolution equations in $3+1$ variables.  Assuming all terms in (\ref{eqn:tdks3d}) are specified, one can use (\ref{eqn:tdks3d}) to simulate molecular systems for which numerical simulation of (\ref{eqn:tdse}) is intractable.

In (\ref{eqn:tdks3d}), the many-body Coulomb interaction between electrons is replaced by known classical Hartree $v^H$ and unknown exchange-correlation $v^{XC}$ single-particle potentials, with the latter incorporating many-body effects. TDDFT is formally an exact theory, as the Runge-Gross and Van Leeuwen theorems proved the  existence of a time-dependent electronic potential and the unique mapping to the time-dependent electron density, which is generated from the KS orbitals of the TDKS equation \citep{runge1984density, vanLeeuwen1999mapping}.

The challenge in TDDFT is to construct $v^{XC}$ potentials that yield an electron density $n$ that is identical to the exact time-dependent many-body electron density generated from the TDSE. Previous work has shown that the unknown $v^{XC}$ formally depends on the initial many-body wave function $\Psi_0$, the initial KS state $\Phi_0$, and the electron density at all points in time $n(r, s<t)$ \citep{maitra2002memory}.  Although the development of $v^{XC}$ for electrons is a very active area of research, almost all $v^{XC}$ make use of the so-called ``adiabatic approximation'' that only takes into account the instantaneous electron density, leading to significant inaccuracies in electron dynamics due to the lack of memory in $v^{XC}$.  \emph{The desire for more accurate electron dynamics leads to a natural question: can we learn $v^{XC}$ from time series data?}  Note that this is an entirely different problem than the problem of learning static, ground state potentials from the exact ground state electron density in \emph{time-independent} density functional theory (DFT) \citep{nagai2018neural, Burke2021review}.

For machine learning of $v^{XC}$ to proceed in the \emph{time-dependent} context (TDDFT), a first obstacle is formulating a tractable learning problem.  In recent work, \cite{suzuki2020machine} works with a \emph{spatially one-dimensional} electron-hydrogen scattering problem.  For this model problem, one can solve (\ref{eqn:tdse}) numerically; from the solution, one can compute the electron density $n(x,t)$ on spatial/temporal grids.  In this problem, we know both the functional form of $v^X$ and that $v^{XC} = v^{X} + v^{C}$.  Furthermore, the one-dimensionality enables one to solve for exact values of $v^{C}$ \citep{elliott2012universal}, again on spatial/temporal grids.  With grid-based values of both $v^{C}$ and $n$, the task of learning $v^C[n]$ becomes a static, supervised learning problem, which \cite{suzuki2020machine} solves using neural network models.  To our knowledge, this is the only prior work on learning $v^{XC}$ for TDDFT.

We revisit the electron-hydrogen scattering model problem and develop methods to learn $v^{C}[n]$ that do not require us to solve for grid-based values of $v^{C}$ beforehand. \emph{In short, we view the $v^C$ functional as a control that guides TDKS propagation.}  We formulate the learning problem as an optimal control problem: find $v^C$ that minimizes the squared error between TDKS electron densities $n$ and reference electron densities $\tilde{n}$.  Implicit in this formulation is the dynamical constraint that electron densities $n$ evolve forward in time via the TDKS equation with the model $v^C$.  The adjoint or costate method is often used to handle constraints of this kind \citep{bryson1975applied, hasdorff1976gradient}.  To our knowledge, the derivations and applications of the adjoint method, \emph{to learn $v^C$ models with memory for the TDKS equation}, are considered here for the first time\footnote{See Section \ref{sect:relationship} in the Appendix for further context.}.  We derive adjoint systems for two settings: (i) to learn pointwise values of $v^C$ on a grid, and (ii) to learn the functional dependence of $v^C$ on the electron density at two points in time.  We apply our methods to train both types of models, and study their training and test performance.  In particular, we train a neural network model of $v^C[n]$ with memory that, when used to solve the TDKS equations for initial conditions outside the training set, yields qualitatively accurate predictions of electron density.


\section{Methods}
\label{sect:methods}
To formulate the problem of learning $v^C$ from time series, we first work in continuous space and time.  Later, to derive numerical algorithms to solve this problem, we discretize.



\paragraph{Continuous Problem.} Define the 1D electron density created from KS orbitals
\begin{equation}
\label{eqn:ndef}
n(x,t) = 2 | \phi(x,t) |^2,
\end{equation}
and the soft-Coulomb external $v^{\text{ext}}$ and interaction $W^{ee}$ potentials
\begin{subequations}
\begin{align}
\label{eqn:vextdef}
 v^{\text{ext}}(x) &= -((x+10)^2+1)^{-1/2}, \\
\label{eqn:Weedef}
W^{ee}(x',x) &= ((x' - x)^2 + 1)^{-1/2}.
\end{align}
\end{subequations}
The potentials (\ref{eqn:vextdef}) and (\ref{eqn:Weedef}) specify that we are working with the spatially one-dimensional electron-hydrogen scattering problem considered by several previous authors.  For this problem, we know that $v^{XC} = v^X + v^C$.  Let $\phi$ and $n$ stand for $\phi(x,t)$ and $n(x,t)$.  Then in one spatial dimension and expressed in atomic units (a.u.), the TDKS system (\ref{eqn:tdks3d}) becomes:
\begin{subequations}
\label{eqn:tdks1d}
\begin{align}
\label{eqn:tdks1devol}
i \partial_t \phi &= -\frac{1}{2} \partial_{xx} \phi + v^{\text{ext}}(x,t) \phi + v^H[n](x,t) \phi + v^X[n](x,t) \phi + v^C[\phi](x,t) \phi, \\
 v^H[n](x,t) &= \int_{x'} W^{ee}(x',x) n(x',t) \, dx', \qquad \qquad v^X[n](x,t) = -\frac{1}{2} v^H[n](x,t).
\end{align}
\end{subequations}

\emph{In (\ref{eqn:tdks1d}), the term that we are trying to learn (e.g., the control) is $v^C[\phi]$.}  Prior first principles work has shown that at time $t$, $v^C$ should depend functionally on the electron density $n(x,s)$ for $s \leq t$, the initial Kohn-Sham state $\phi(x,0)$ and the initial Schr\"odinger wave function $\Psi(x,0)$ \citep{maitra2002memory, wagner2012exact}.  In this work, we ignore the dependence of $v^C$ on the initial states $\phi(x,0)$ and $\Psi(x,0)$, and focus on modeling the dependence on present and past electron densities.  By (\ref{eqn:ndef}), dependence on $n$ is equivalent to a particular type of dependence on $\phi$; we use the notation $v^C[\phi]$ to refer to models that depend  on $\phi$ either directly or through $n$.

\emph{For the sake of intuition, let us formulate the control problem in continuous time and space.}  Assume that for $t \in [0, T]$, we have access to a reference electron density trajectory $\tilde{n}(x,t)$.  Suppose that our model $v^C[\phi; \btheta]$ is parameterized by $\btheta$.  Then we seek to minimize the squared loss
\begin{equation}
\label{eqn:l2loss}
\mathcal{J} (\btheta) = \frac{1}{2} \int_{x=-\infty}^{\infty} \int_{t=0}^T ( n(x,t) - \tilde{n}(x,t) )^2 \, dt \, dx,
\end{equation}
subject to the constraint that $n(x,t)$ is computed via (\ref{eqn:ndef}), with $\phi(x,t)$ evolving on the interval $0 \leq t \leq T$ according to the TDKS system (\ref{eqn:tdks1d}).  In this TDKS system, we identify $v^C$ with our model $v^C[\phi; \btheta]$.  \emph{In short, we seek $\btheta$ such that the resulting $v^C[\phi; \btheta]$ functional guides the TDKS system to yield a solution $\phi(x,t)$ such that $n = 2 |\phi|^2$ matches the reference trajectory $\tilde{n}(x,t)$.}

\paragraph{Direct and Adjoint Methods.}  In a \emph{direct method} to minimize the loss (\ref{eqn:l2loss}), we compute gradients by applying $\nabla_{\btheta}$ to both sides of (\ref{eqn:l2loss}).  This will yield an expression for $\nabla_{\btheta} \mathcal{J}$ that involves $\nabla_{\btheta} \phi$.  To compute this latter quantity, we numerically solve an evolution equation derived by taking $\nabla_{\btheta}$ of both sides of (\ref{eqn:tdks1devol}).  At each iteration of our gradient-based optimizer, we would carry out this procedure to compute $\nabla_{\btheta} \mathcal{J}$, which is then used to update $\btheta$.  In practice, this direct method suffers from one major problem: if we discretize $\phi$ in space using $J+1$ grid points, and if $\btheta$ has dimension $B$, then at each point in time, $\nabla_{\btheta} \phi$ will have dimension $(J+1)B$.  In our work, $B$ can exceed $10^7$, while $J \geq 600$ is required for sufficient spatial accuracy.  Solving the evolution equation for $\nabla_{\btheta} \phi$ in $(J+1)B$-dimensional space thus incurs huge computational expense at each optimization step.

In this paper, we pursue the \emph{adjoint method}, which enables us to compute all required gradients without computing or even storing any $(J+1)B$-dimensional objects, thus dramatically reducing computational costs relative to the direct method. Within the space of adjoint methods, there are two broad approaches: (i) to use the continuous-time loss and constraints to derive differential equations for continuous-time adjoint variables, and (ii) to first discretize the loss and constraints, and then derive numerical schemes for discrete-time adjoint variables.  In approach (i), we must still discretize the adjoint differential equations in order to solve them; the choice of discretization can lead to subtle issues \citep{SanzSerna}.  We choose approach (ii) for its relative simplicity.

In the discrete adjoint method, we incorporate a discretized version of the dynamical system (\ref{eqn:tdks1d}) as a constraint using time-dependent Lagrange multipliers $\blambda(t)$.  In this approach, we derive and numerically solve a backward-in-time evolution equation for $\blambda(t)$, from which we compute required gradients.  Importantly, $\blambda(t)$ has the same dimension as the state variables $\bphi(t)$ defined below; in our implementation, both quantities are $(J+1)$-dimensional.  We obtain the gradients of the discretized loss at a computational cost that is proportional to that of computing the loss itself.

\paragraph{Discretized Problem.} To keep this paper focused on the learning/control problem, we have moved details of the numerical solution of the TDKS system (\ref{eqn:tdks1d}) to Section \ref{sect:tdksforward} of the Appendix. Here we include only the most important concepts.  First, we discretize the Kohn-Sham state by introducing $\bphi(t_k) = [\phi(x_{0}, t_k), \ldots, \phi(x_J, t_k)]^T$.  The spatial domain is $x \in [L_{\text{min}}, L_{\text{max}}]$. With $\Delta x = (L_{\text{max}} - L_{\text{min}})/J > 0$, our spatial grid is $x_j = L_{\text{min}} + j \Delta x$.  Our temporal grid is $t_k = k \Delta t$, with $\Delta t = T/K$.  The positive integers $J$ and $K$ are user-defined parameters that control the accuracy of the discretization.  Second, by applying finite differences, Simpson's quadrature rule, and operator splitting, we can derive the following evolution equation for the discretized state $\bphi$ defined above:
\begin{equation}
\label{eqn:tdksDISC}
\bphi(t_{k+1}) = \exp(-i \mathcal{K} \Delta t/2) \exp(-i V(\bphi(t_k), \bv^{C}_k) \Delta t) \exp(-i \mathcal{K} \Delta t/2) \bphi(t_k).
\end{equation}
Here $\mathcal{K}$ is a constant $(J+1) \times (J+1)$ matrix, while $V$ is a diagonal $(J+1) \times (J+1)$ matrix that depends functionally on both the state $\bphi$ and on $\bv^{C}$, our spatially discretized model of the correlation potential $v^C$ from (\ref{eqn:tdks1d}).  Detailed descriptions of $\mathcal{K}$ and $V$ are provided in Section \ref{sect:tdksforward}.

Evolving $\bphi$ according to (\ref{eqn:tdksDISC}) generates a numerical approximation to the solution $\phi(x,t)$ of (\ref{eqn:tdks1d}).  This approximation has a truncation error of $O(\Delta t^2)$ in time and $O(\Delta x^4)$ in space.

\paragraph{First Adjoint Method: Learning $v^C$ Pointwise.} Assume we have access to observed values of electron density on the grid---we denote these observed or reference values by  $\tilde{n}(x_j,t_k)$.  The first problem we consider is to learn $v^C(x_j, t_k)$ on the same grid.  Suppose we start from an initial condition $\bphi(0)$ and an estimate $\bv^{C}$.  We iterate (\ref{eqn:tdksDISC}) forward in time and obtain a trajectory $\bphi(t_k)$ for $0 \leq k \leq K$.  We then form $n(x_j,t_k) = |\phi(x_j,t_k)|^2$.  In this subsection, $\phi$ and $n$ are the predicted wave function and density when we use the estimated correlation potential $\bv^{C}$.  Let $\mathcal{P}_\mathcal{K} = \exp(-i \mathcal{K} \Delta t/2)$ and abbreviate $\bphi_k = \bphi(t_k)$, $\bv^C_k = \bv^C(t_k)$.  Define the discrete-time propagator
\begin{equation}
\label{eqn:discprop}
\bF_{\Delta t}(\bphi, \bv^{C}) = \mathcal{P}_\mathcal{K} \exp(-i V(\bphi, \bv^C) \Delta t) \mathcal{P}_\mathcal{K} \bphi,
\end{equation}
so that (\ref{eqn:tdksDISC}) can be written as the discrete-time system $\bphi_{k+1} = \bF_{\Delta t}(\bphi_k, \bv^{C}_k)$.  Both sides of this system are complex-valued.  In order to form a real-valued Lagrangian and take real variations, we split both $\bphi$ and $\bF$ into real and imaginary parts: $\bphi = \bphi^R + i \bphi^I$ and $\bF_{\Delta t} = \bF_{\Delta t}^R + i \bF_{\Delta t}^I$.  Superscript $R$ and $I$ denote, respectively, the real and imaginary parts of a complex quantity.  Let the uppercase $\bPhi$, $\bLambda$, and $\bV^{C}$ denote the collections of all corresponding lowercase $\bphi_k$, $\blambda_k$, and $\bv^{C}_k$ for all $k$.   Then we form a real-variable Lagrangian that consists of the discretized squared loss with the constraint that $\bphi$ evolves via (\ref{eqn:tdksDISC}).
\begin{multline}
\label{eqn:lag}
\mathscr{L}(\bPhi^R, \bPhi^I, \bLambda^R, \bLambda^I, \bv^{C}) = \frac{1}{2} \sum_{k=0}^{K} \sum_{j=0}^{J} ( 2 \phi^R(x_j, t_k)^2 + 2 \phi^I(x_j, t_k)^2 - \tilde{n}(x_j, t_k) )^2 \\
- \sum_{k=0}^{K-1} [\blambda_{k+1}^R]^T ( \bphi_{k+1}^R  - \bF^R_{\Delta t}(\bphi_k^R, \bphi_k^I, \bv^{C}_k) ) +  [\blambda_{k+1}^I]^T ( \bphi_{k+1}^I  - \bF^I_{\Delta t}(\bphi_k^R, \bphi_k^I, \bv^{C}_k)  ).
\end{multline}
 Setting $\delta \mathscr{L} = 0$ for all variations $\delta \bphi_k^R$ and $\delta \bphi_k^I$ for $k \geq 1$, we obtain
\begin{subequations}
\label{eqn:adjsys}
\begin{align}
\label{eqn:adjfin}
\blambda_K &= 4 \left[ (2 |\bphi_K|^2 - \tilde{\bn}_K) \circ \bphi_K \right] \\
\label{eqn:adjeqn}
\begin{bmatrix} \blambda_{k}^R  \\  \blambda_{k}^I \end{bmatrix}^T &= 4  (2 |\bphi_k|^2 - \tilde{\bn}_k) \circ \begin{bmatrix} \bphi_k^R \\ \bphi_k^I \end{bmatrix}^T + 
\begin{bmatrix} \blambda_{k+1}^R  \\  \blambda_{k+1}^I \end{bmatrix}^T \bJ_{\bphi} \bF_{\Delta t} (\bphi_k^R, \bphi_k^I, \bv^{C}_k).
\end{align}
\end{subequations}
Here $\bJ_{\bphi} \bF_{\Delta t}$ denotes the Jacobian of $\bF$ with respect to $\bphi$.  We use (\ref{eqn:adjfin}) as a final condition and iterate (\ref{eqn:adjeqn}) backward in time for $k = K-1, \ldots, 1$.  Having computed $\bLambda$ from (\ref{eqn:adjsys}), we return to (\ref{eqn:lag}) and compute the gradient with respect to $\bv^{C}_{\ell}$:
\begin{equation}
\label{eqn:nablavc}
\nabla_{\bv^{C}_{\ell}} \mathscr{L} = \begin{bmatrix} \blambda_{\ell+1}^R  \\  \blambda_{\ell+1}^I \end{bmatrix}^T  \nabla_{\bv^{C}_{\ell}} \begin{bmatrix} \bF_{\Delta t}^R  \\
\bF_{\Delta t}^I \end{bmatrix} (\bphi_{\ell}^R, \bphi_{\ell}^I, \bv^{C}_{\ell}).
\end{equation}
Given a candidate $\bv^{C}$, we solve the forward problem to obtain $\bPhi$.  We then solve the adjoint system to obtain $\bLambda$.  This provides everything required to evaluate (\ref{eqn:nablavc}) for each $\ell$.  The variations, the block matrix form of the Jacobian $\bJ_{\bphi} \bF_{\Delta t}$, and the gradients of the discrete-time propagator $\bF$ can be found in Sections \ref{sect:adjderiv} and \ref{sect:TDKSgrads} of the preprint Appendix.

\paragraph{Second Adjoint Method: Learning $v^C$ Functionals.}  Here we rederive the adjoint method to enable learning the \emph{functional dependence} of $v^C[\phi](x,t)$ on $\phi(x,t)$ and $\phi(x,t - \Delta t)$.  We take as our model
$v^C[\phi] = v^C(\bphi, \bphi'; \btheta)$.
The parameters $\btheta$ determine a particular functional dependence of $v^C$ on the present and previous Kohn-Sham states $\bphi$ and $\bphi'$.  At spatial grid location $x_j$ and time $t_k$, the model $v^C$ is
\begin{equation}
\label{eqn:vcmemorymodel2}
v^C[\phi](x_j, t_k) = [ \bv^C(\bphi_{k}, \bphi_{k-1}; \btheta) ]_j.
\end{equation}
In short, we intend $\bphi'$ to be the Kohn-Sham state at the time step \emph{prior} to the time step that corresponds to $\bphi$.  Our goal is to learn $\btheta$.  This requires redefining the following quantities:
\begin{align*}
V(\bphi; \btheta) &= \operatorname{diag}( \mathbf{v}(\bphi,\bphi'; \btheta)) \\
\bv(\bphi, \bphi'; \btheta) &=-((\bx + 10)^2 + 1)^{-1/2} + W  ( |\bphi|^2 \circ \bw ) + \bv^{C}(\bphi,\bphi'; \btheta) \\
\bF_{\Delta t}(\bphi,\bphi'; \btheta) &= \mathcal{P}_\mathcal{K} \exp(-iV(\bphi,\bphi'; \btheta) \Delta t) \mathcal{P}_\mathcal{K} \bphi.
\end{align*}
The Lagrangian still has the form of an objective function together with a dynamical constraint:
\begin{multline}
\label{eqn:newlag}
\mathscr{L}(\bPhi^R, \bPhi^I, \bLambda^R, \bLambda^I, \btheta) = \frac{1}{2} \sum_{k=0}^{K} \sum_{j=0}^{J} ( 2 \phi^R(x_j, t_k)^2 + 2 \phi^I(x_j, t_k)^2 - \tilde{n}(x_j, t_k) )^2 - \sum_{k=1}^{K-1} [\blambda_{k+1}^R]^T \\
 ( \bphi_{k+1}^R  - \bF^R_{\Delta t}(\bphi_k^R, \bphi_{k-1}^R, \bphi_k^I, \bphi_{k-1}^I; \btheta) ) 
+  [\blambda_{k+1}^I]^T ( \bphi_{k+1}^I  - \bF^I_{\Delta t}(\bphi_k^R, \bphi_{k-1}^R, \bphi_k^I, \bphi_{k-1}^I; \btheta)  )
\end{multline}
Setting $\delta \mathscr{L} = 0$ for all variations $\bphi^R_k$ and $\bphi^I_k$ for $k \geq 1$, we obtain the following adjoint system:
\begin{subequations}
\label{eqn:newadj}
\begin{align}
\label{eqn:lambK}
\blambda_K &= 4 \left[ (2 |\bphi_K|^2 - \tilde{\bn}_K) \circ \bphi_K \right] \\
\label{eqn:lambKm1}
\blambda_{K-1} &= 4 \left[ (2 |\bphi_{K-1}|^2 - \tilde{\bn}_{K-1}) \circ \bphi_{K-1} \right] \\
 &\qquad + [\blambda_{K}^R]^T \nabla_{\bphi} \bF_{\Delta t}^R(\bphi_K, \bphi_{K-1}; \btheta) + [\blambda_{K}^I]^T \nabla_{\bphi} \bF_{\Delta t}^I(\bphi_K, \bphi_{K-1}; \btheta) \nonumber \\
\label{eqn:newadjiter}
\begin{bmatrix} \blambda_{k}^R  \\  \blambda_{k}^I \end{bmatrix}^T \! \! \! \! &= 4  (2 |\bphi_k|^2 \! \! - \!  \tilde{\bn}_k) \! \circ  \! \begin{bmatrix} \bphi_k^R \\ \bphi_k^I \end{bmatrix}^T \! \! \! \! \! +  \! \begin{bmatrix} \blambda_{k+1}^R  \\  \blambda_{k+1}^I \end{bmatrix}^T \! \! \! \bJ_{\bphi} \bF_{\Delta t} (\bphi_k, \bphi_{k-1}; \btheta)  + \begin{bmatrix} \blambda_{k+2}^R  \\ \blambda_{k+2}^I \end{bmatrix}^T \! \! \! \bJ_{\bphi'} \bF_{\Delta t} (\bphi_{k+1}, \bphi_k; \btheta) 
\end{align}
\end{subequations}
The key difference between (\ref{eqn:newadjiter}) and (\ref{eqn:adjeqn}) is that the right-hand side of (\ref{eqn:newadjiter}) involves $\blambda$ at two points in time.  The adjoint system is now a linear delay difference equation with time-dependent coefficients.  Additionally, the derivatives of $F_{\Delta t}$ needed to evaluate (\ref{eqn:newadj}-\ref{eqn:nablatheta}) are different---see Section \ref{sect:TDKSgrads} of the preprint Appendix. For a candidate value of $\btheta$, we solve the forward problem to obtain $\bphi$ on our spatial and temporal grid.  Then, to compute gradients, we begin with the final conditions (\ref{eqn:lambK}-\ref{eqn:lambKm1}) and iterate (\ref{eqn:newadjiter}) backwards in time from $k = K-2$ to $k=1$.  Having solved the adjoint system, we compute the gradient of $\mathscr{L}$ with respect to $\btheta$ via
\begin{equation}
\label{eqn:nablatheta}
\nabla_{\btheta} \mathscr{L} = \sum_{k=1}^{K-1}  \begin{bmatrix} \blambda_{k+1}^R  \\  \blambda_{k+1}^I \end{bmatrix}^T  \nabla_{\btheta} \begin{bmatrix} \bF_{\Delta t}^R  \\
\bF_{\Delta t}^I \end{bmatrix} (\bphi_k, \bphi_{k-1}; \btheta).
\end{equation}

\begin{figure}
\centering \includegraphics[width=5in]{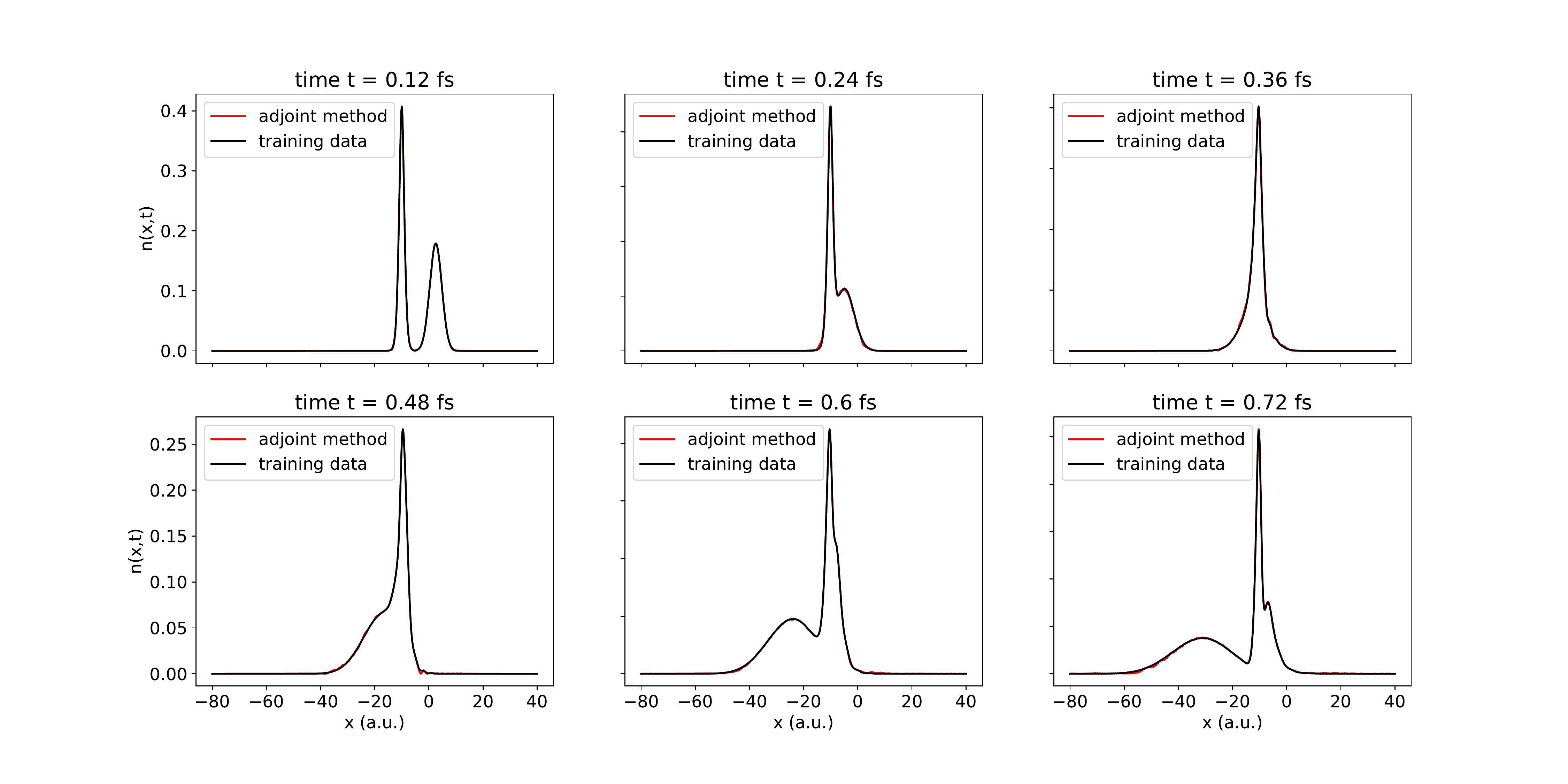}
\caption{Training results for the problem of learning pointwise values of $\bV^C$ on a grid consisting of $K = 30000$ points in time and $J = 600$ points in space.  The adjoint method succeeds in producing $\bV^C$ values that yield TDKS solutions such that the corresponding electron densities (red) match those computed from the 2D Schr\"odinger equation (black).}
\label{fig:vCgridtrain}
\end{figure}

\section{Modeling and Implementation Details}
\label{sect:deets}
\paragraph{Modeling Correlation Functionals.}
In this work, all models of the form (\ref{eqn:vcmemorymodel2}) consist of dense, feedforward neural networks.  For models of the form $\bv^C(\bphi, \bphi'; \btheta)$, we treat the real and imaginary parts of $\bphi$ and $\bphi'$ as real vectors each of length $J+1$.  Hence for $J = 600$, we have an input layer of size $4(J+1)$.  We follow this with three hidden layers each with $256$ units and a scaled exponential linear unit  activation function \citep{selupaper}. The output layer has $J+1$ units to correspond to the vector-valued output $\bv^C$.  For models in which $\bv^C$ depend on $\bn$ and $\bn'$, we take the real and imaginary parts of $\bphi$ and $\bphi'$ as inputs and use them to immediately compute $\bn$ and $\bn'$, which we then concatenate and feed into an input layer with $2(J+1)$ units.  The remainder of the network is as above.   We started with smaller networks (fewer layers, less units per layer) and increased the network size until we obtained reasonable training results; no other architecture search or hyperparameter tuning was carried out.   We experimented with other activation functions and convolutional layers---none of these models produced satisfactory results during training.  

\paragraph{Generation of Training Data.}
To generate training data, we solve (\ref{eqn:tdse}) for a model system consisting of $N=2$ electrons: a one-dimensional electron scattering off a one-dimensional hydrogen atom.  Hence (\ref{eqn:tdse}) becomes a partial differential equation (PDE) for a wave function $\Psi(x_1,x_2,t)$.  We discretize this PDE using finite differences on an equispaced grid in $(x_1, x_2)$ space with $J=1201$ points along each axis.  Here $-80 \leq x_1, x_2 \leq 40$, so that $\Delta x = 0.1$.  After discretizing the kinetic and potential operators in space, we propagate forward in time until $T = 0.72$ fs, using second-order operator splitting with $\Delta t = 2.4 \times 10^{-5}$ fs (or, in a.u., $\Delta t \approx 9.99219 \times 10^{-4}$).  Note that this is $1/100$-th the time step used by \cite{suzuki2020machine}.  For further details, consult Section \ref{sect:schroforward}.  After discretization, the wave function $\Psi(x_1,x_2,t)$ at time step $k$ is a complex vector $\bpsi_k$ of dimension $(J+1)^2$.  For the initial vector $\bpsi_0$, we follow \cite{suzuki2020machine} and use a Gaussian wave packet that represents an electron initially centered at $x=10$ a.u., approaching the H-atom localized at $x=-10$ a.u., with momentum $p$.  We generate training/test data by numerically solving the Schr\"odinger system for initial conditions with $p \in \{-1.0, -1.2, -1.4, -1.5, -1.6, -1.8\}$.  From the resulting time series of wave functions, we compute the time-dependent one-electron density $n(x,t)$; below, \emph{we refer to this as the TDSE electron density}.

\section{Results}
\label{sect:results}

\begin{figure}
\centering \includegraphics[width=5in]{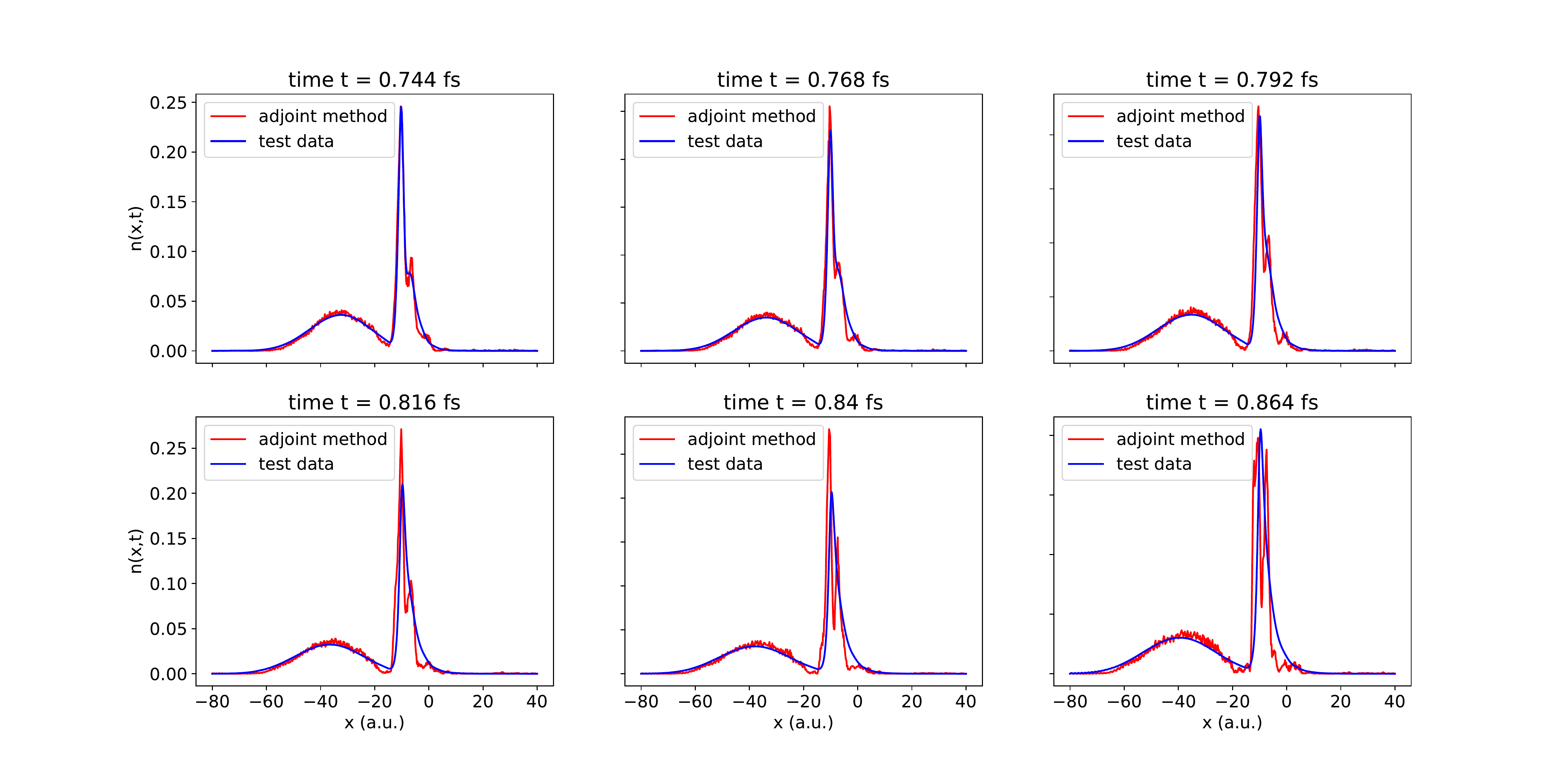}
\caption{We use $300$ time steps (corresponding to $0.72$ fs) of the $p=-1.5$ data together with the adjoint method to train a neural network model of $v^C$ that depends on the current and previous $\phi$.  Using the learned $v^C$, we propagate (\ref{eqn:tdks1d}) for $60$ additional time steps and plot the test set results (in red) against the reference electron density (in blue).}
\label{fig:vCp15TEST}
\end{figure}

\paragraph{Pointwise Results.} Our first result concerns learning the pointwise values of $\bV^C$.  Here we use the same fine time step $\Delta t = 2.4 \times 10^{-5}$ used to generate the training data.  However, we increase $\Delta x$ by a factor of $2$, taking $J = 600$ and sampling the initial condition $\bphi_0$ at every other grid point.  We retain this subsampling in space in all training sets/results that follow. Still, our unknown $\bV^C$ consists of a total of $30000 \cdot 601$ values.  

We learn $\bV^C$ by optimizing an objective function that consists of the first line of (\ref{eqn:lag}) together with a regularization term. The regularization consists of a finite-difference approximation of $\mu \sum_{k} \sum_{j} ( \partial_x v^C(x_j, t_k) )^2$, with $\mu = 10^{-5}$.  This regularization is analogous to the $\int (f'')^2 dx$ penalty used in smoothing splines \citep{ESLII}.  We penalize the square of the first (rather than second) derivative as we find this is sufficient to smooth $v^C$ in space.  The precise value of $\mu$ is unimportant; taking $\mu \in [10^{-6}, 10^{-4}]$ yields similar results. For training data, we use only the TDSE one-electron densities computed from the $p=-1.5$ initial condition. To optimize, we use the quasi-Newton L-BFGS-B method, with gradients $\nabla_{\bV^C} \mathscr{L}$ computed via the procedure described just below (\ref{eqn:nablavc}).  We initialize the optimizer with $\bV^C \equiv 0$ and use default tolerances of $10^{-6}$.

In Figure \ref{fig:vCgridtrain}, we present the results of this approach.  Each panel shows a snapshot of both the training electron density (in black, computed from TDSE data) and the electron density $n = 2 |\phi|^2$ (in red) obtained by solving TDKS (\ref{eqn:tdks1d}) using the learned $\bV^C$ values. Note the close quantitative agreement between the black and red curves.  The overall mean-squared error (MSE) across all points in space and time is $2.035 \times 10^{-6}$.  Note that no exact $\bV^C$ data was used; the learned $\bV^C$ does not match the exact $\bV^C$ quantitatively, but does have some of the same qualitative features.

This problem suits the adjoint method well: regardless of the dimensionality of $\bV^C$, the dimensionality of the adjoint system is the same as that of the discretized TDKS system.  Note that, for this one-dimensional TDKS problem (\ref{eqn:tdks1d}), it is possible to solve for $\bV^C$ on a grid \citep{elliott2012universal}. If we encounter solutions of \emph{higher-dimensional, multi-electron} ($d \geq 2$ and $N \geq 2$) Schr\"odinger systems from which we seek to learn $\bV^C$, we will not be able to employ an exact procedure.  In this case, the adjoint-based method may yield numerical values $\bV^C$, with which we can pursue supervised learning of a functional from electron densities $\bn$ to correlation potentials $\bV^C$.

\begin{figure}
\centering \includegraphics[width=5in]{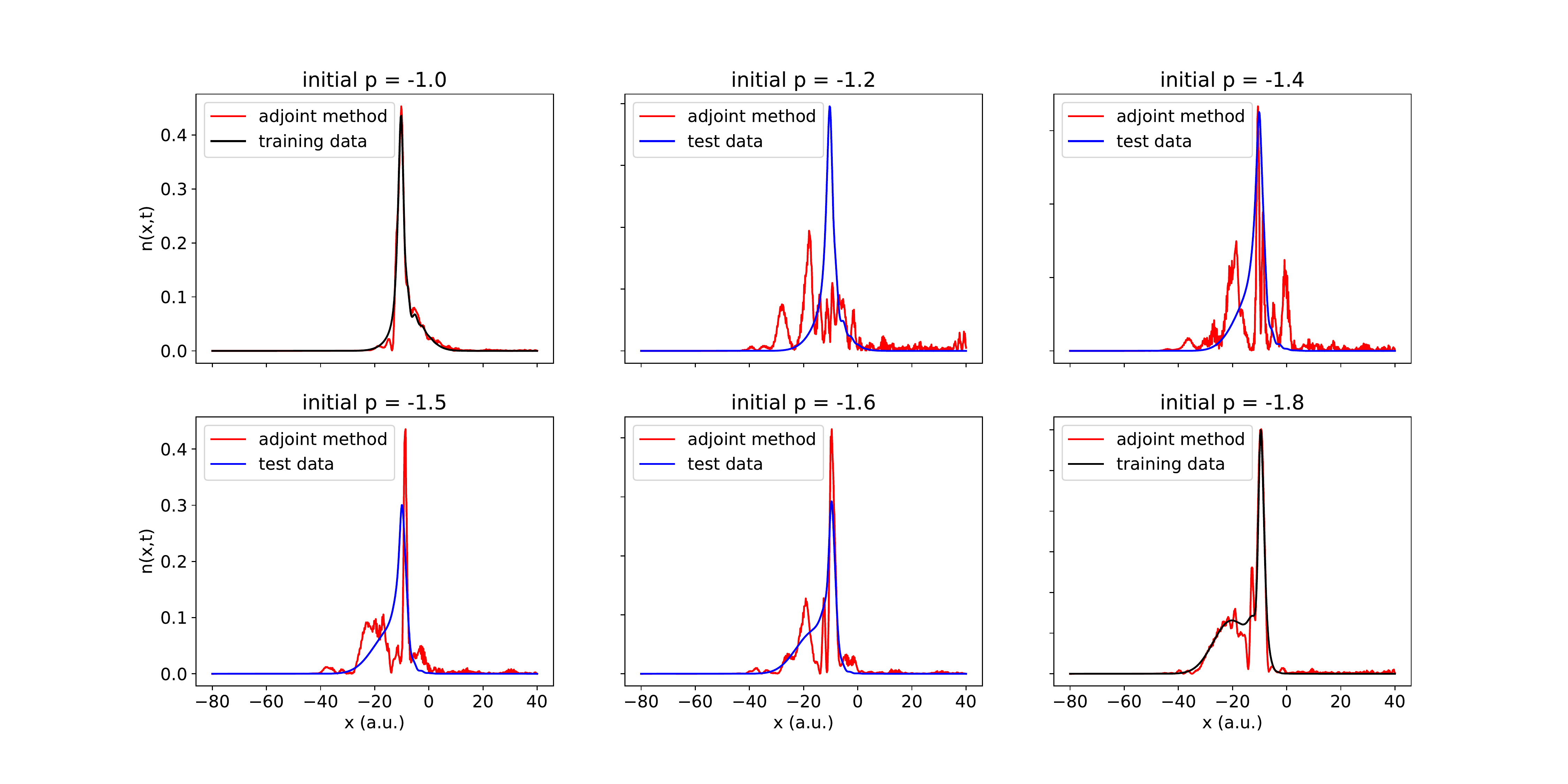}
\caption{We plot training and test results at time $t = 0.432$ fs for the adjoint method, applied to estimating neural network models $v^C[\phi]$ that, at time $t$, depend on both $\phi(x,t)$ and $\phi(x, t-\Delta t)$.  Propagating  TDKS (\ref{eqn:tdks1d}) with the learned $v^C$ yields the red curves.}
\label{fig:vCphimemory}
\end{figure}

\paragraph{Functional Results.} Next we present results in which we learn $v^C$ functionals.  In preliminary work, we sought to model $v^C[\phi](x,t)$ as purely a function of $\phi(x,t)$, a model without memory.  These models did not yield satisfactory training set results, and hence were abandoned.  We focus first on models $v^C[\phi](x,t)$ that allow for arbitrary dependence on the real and imaginary parts of both $\phi(x,t)$ and $\phi(x,t-\Delta t)$.   The TDDFT literature emphasizes that $v^C$ should depend on $\phi$ through present/past electron densities $n$, where $n = 2 | \phi|^2$.  How important is it to incorporate such physics-based constraints into our $v^C$ model?  Let us see how well a direct neural network model of $v^C[\phi]$ captures the dynamics.  The input layer is of dimension $4(J+1)$---see Section \ref{sect:deets}.

To train such a model, we again apply the L-BFGS-B optimizer with objective function given by the first line of (\ref{eqn:newlag}) and gradients computed with the adjoint system (\ref{eqn:newadj}-\ref{eqn:nablatheta}).  We initialize neural network parameters $\btheta$ by sampling a mean-zero normal distribution with standard deviation $\sigma = 0.01$.  For training data, we subsample the $p=-1.5$ TDSE electron density time series by a factor of $100$ in time, so that $\Delta t = 2.4 \times 10^{-3}$ fs and the entire training trajectory consists of $K=301$ time steps.  We retain this time step in all training sets and results that follow.

We omit the training set results here as they show excellent agreement between training and model-predicted electron densities---see Section \ref{sect:trainresults}.  The overall training set mean-squared error (MSE) is $7.668 \times 10^{-6}$.  In Figure \ref{fig:vCp15TEST}, we display test set results obtained by propagating for $60$ additional time steps beyond the end of the training data.  On this test set, we see close quantitative agreement near $t=0.72$ fs, which slowly degrades.  Still, the learned $v^C$ leads to TDKS electron densities that capture essential features of the reference trajectory.  Note that no regularization was used during training of the $v^C$ functional, leading to a learned $v^C$ that is not particularly smooth in space.  We hypothesize that, with careful and perhaps physically motivated regularization, the learned $v^C$ will yield improved test set results over longer time intervals.

In the next set of results, we retrain our model using TDSE electron densities with initial momenta equal to $p=-1.0$ and $p=-1.8$.  We train two models: a $v^C[\phi](x,t)$ model that depends on $\phi$  at times $t$ and $t - \Delta t$, and a $v^C[n](x,t)$ model that depends on $n$ at times $t$ and $t - \Delta t$.  This latter model incorporates the physics-based constraints mentioned above.  We view the $v^C[n]$ model as more constrained because its the first hidden layer can depend on $\phi(x,t)$ and $\phi(x,t-\Delta t)$ \emph{only through} the electron densities $n(x,t)$ and $n(x,t-\Delta t)$. We keep all other details of training the same.  The final training set MSE values are $4.645 \times 10^{-5}$ for the $v^C[\phi]$ model and $8.098 \times 10^{-5}$ for the $v^C[n]$ model.

\begin{figure}
\centering \includegraphics[width=5in]{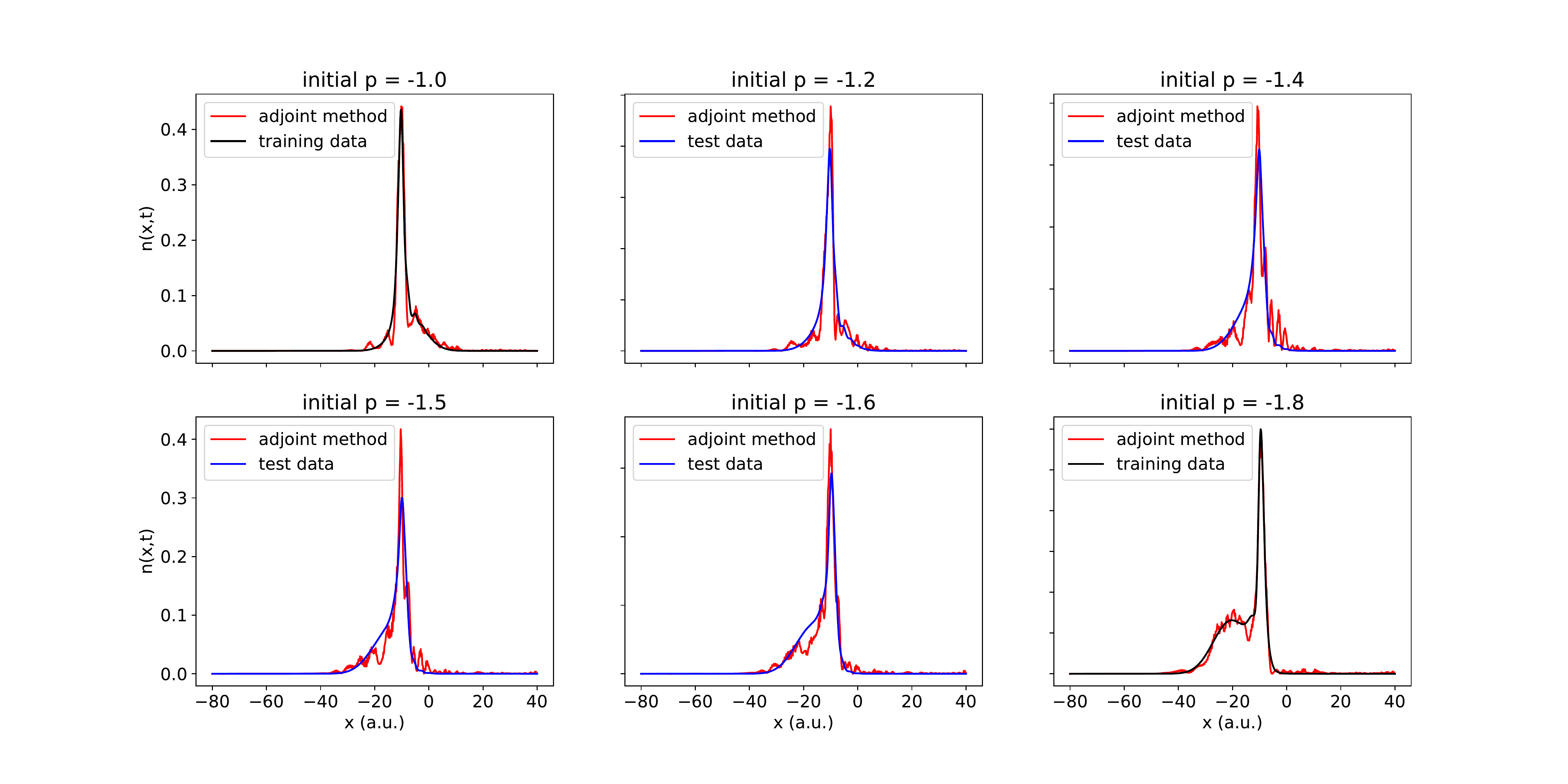}
\caption{We plot training and test results at time $t = 0.432$ fs for the adjoint method, applied to estimating a neural network model $v^C[n]$ that, at time $t$, depends on both $n(x,t)$ and $n(x, t-\Delta t)$.  Propagating TDKS (\ref{eqn:tdks1d}) with the learned $v^C[n]$ yields the red curves.}
\label{fig:vCnvecmemory}
\end{figure}

In Figures \ref{fig:vCphimemory} and \ref{fig:vCnvecmemory}, we plot both training and test set results for these models.  Here we have chosen a particular time ($t = 0.432$ fs) and plotted the electron density at this time for six different trajectories, each with a different initial momentum $p$.  We have chosen this time to highlight the large, obvious differences between the $p=-1.0$ and $p=-1.8$ curves.  The $p=-1.0$ and $p=-1.8$ panels contain training set results; here the TDKS electron densities (in red, produced using the learned $v^C$) lie closer to the ground truth TDSE electron densities (in black).

Note that, despite the greater freedom enjoyed by the $v^C[\phi]$  model, its generalization to trajectories \emph{outside the training set} ($-1.2 \leq p \leq -1.6$) is noticeably worse than that of the more constrained $v^C[n]$ model.    In fact, the $v^C[n]$ model's results (Figure \ref{fig:vCnvecmemory}, in red) show broad qualitative agreement with the test set TDSE curves (in blue).  The test set MSE values are $9.363 \times 10^{-4}$ for the $v^C[\phi]$ model and $2.482 \times 10^{-4}$ for the $v^C[n]$ model.  Overall, these results support the view that $v^C$ should depend on $\phi$ through $n$.  Again, we hypothesize that if we were to filter out short-wavelength oscillations in the electron density---perhaps by regularizing the $v^C[n]$ model or by training on a larger set of trajectories---the agreement could be improved.  

\paragraph{Conclusion.} For a low-dimensional model problem, we have developed adjoint-based methods to learn the correlation potential $v^C$ using data from TDSE simulations.  The adjoint method can be used to \emph{directly train} $v^C[n]$ models, sidestepping the need for either exact $v^C$ values or density-to-potential inversion.    Our work provides a foundation for learning models that depend on present and past snapshots of the electron density.  We find that our trained $v^C[n]$ models (with memory) generalize well to trajectories outside the training set.  Further improvements to the model may be possible, e.g., by incorporating known physics in the form of model constraints.  Overall, the results show the promise of learning $v^C$ via TDKS-constrained optimization.

\newpage

\acks{This work was supported by the U.S. Department of Energy, Office of Science, Basic Energy Sciences under Award Number DE-SC0020203. This research used resources of the National Energy Research Scientific Computing Center (NERSC), a U.S. Department of Energy Office of Science User Facility located at Lawrence Berkeley National Laboratory, operated under Contract No. DE-AC02-05CH11231 using NERSC award BES-m2530 for 2021.
We acknowledge computational time on the Pinnacles cluster at UC Merced (supported by NSF OAC-2019144).  We also acknowledge computational time on the Nautilus cluster, supported by the Pacific Research Platform (NSF ACI-1541349), CHASE-CI (NSF CNS-1730158), and Towards a National Research Platform (NSF OAC-1826967). Additional funding for Nautilus has been supplied by the University of California Office of the President.}

\newpage
 \pagenumbering{gobble}

\section{Appendix}
\label{sect:Appendix}

\subsection{Solving the Spatially One-Dimensional TDKS System}
\label{sect:tdksforward}
To solve (\ref{eqn:tdks1d}), we use a finite-difference discretization of (\ref{eqn:tdks1d}) on the spatial domain $x \in [L_{\text{min}}, L_{\text{max}}]$ and temporal domain $t \in [0, T]$.  Fix spatial and temporal grid spacings $\Delta x > 0$ and $\Delta t > 0$; let $x_j = L_{\text{min}} + j \Delta x$ and $t_k = k \Delta t$.  Then our spatial grid is $\{x_j\}_{j=0}^{j=J}$ with $J \Delta x = L_{\text{max}} - L_{\text{min}}$, and our temporal grid is $\{t_k\}_{k=0}^K$ with $K \Delta t = T$.

Suppose the correlation functional has been specified in one of two ways: (i) for a particular trajectory, we have access to the \emph{values} of $v^C(x_j, t_k)$ at all spatial and temporal grid points $x_j$ and $t_k$, or (ii) we have a model $v^C[\phi]$ that takes as input $\phi(x,s)$ for $s \leq t_k$ and produces as output $v^C(x_j, t_k)$ for all $j$.  Then, given an initial condition $\phi(x,0)$, the \emph{forward problem} is to solve (\ref{eqn:tdks1d}) numerically on the grids defined above.

Our first step is to discretize (\ref{eqn:tdks1d}) in space.  Let $\bphi(t)$ be the $(J+1) \times 1$ column vector
\begin{equation}
\bphi(t) = [\phi(x_{0}, t), \ldots, \phi(x_J, t)]^T,
\end{equation}
where $^T$ denotes transpose.  We use the integers $0, \ldots, J$ to index this vector, so that $[\bphi(t)]_{j} = \phi(x_j, t)$.  We discretize $\partial_{xx}$ with a fourth-order Laplacian matrix $\Delta$, defined in (\ref{eqn:discreteLaplacian}),  such that
\begin{equation}
\label{eqn:laplacian}
\partial_{xx} \phi(x,t) \bigr|_{x=x_j} = [\Delta \bphi(t)]_{j} + O(\Delta x^4).
\end{equation}
Besides the $\partial_{xx}$ term, the remaining terms on the right-hand side of (\ref{eqn:tdks1d}) result in a diagonal matrix multiplied by $\bphi$.  The only term that requires further numerical approximation is the integral term.  For this purpose, we define the symmetric matrix $W_{j,j'} = ((x_{j'} - x_j)^2 + 1)^{-1/2} \Delta x$.
Let $\circ$ denote the entry-wise product of vectors and let $|\bphi(t)|^2$ be the vector of entry-wise magnitudes $|\phi(x_j,t)|^2$.  With quadrature weights $w_{j}$ given by Simpson's rule, the integral at $x=x_j$ becomes
\begin{align}
\int_{x'} ((x' - x_j)^2 + 1)^{-1/2} | \phi(x',t)|^2 dx' &= \sum_{j'} W_{j,j'} \phi(x_{j'},t) \phi^\ast(x_{j'},t) w_{j'} + O(\Delta x^{4}) \nonumber \\
\label{eqn:quadrature}
 &= \left[ W ( |\bphi(t)|^2 \circ \bw ) \right]_{j} + O(\Delta x^{4})
\end{align}
Let $\mathcal{K} = -(1/2) \Delta$ be the spatially discretized kinetic operator, with $\Delta$ defined by the following fourth-order discrete Laplacian:
\begin{equation}
\label{eqn:discreteLaplacian}
\Delta = \frac{1}{12 \Delta x^2} \begin{bmatrix}
-30 & 16 & -1 & &  \\
16 & \ddots & \ddots & \ddots & \\
-1 & \ddots & \ddots & \ddots & \ddots \\
& \ddots & \ddots & \ddots & \ddots & -1 \\
& & \ddots & \ddots & \ddots &  16 \\
& & & -1 & 16 & -30
\end{bmatrix}.
\end{equation}
Evaluating both sides of (\ref{eqn:tdks1d}) at $x=x_j$ for all $j$ at once, we arrive at the following nonlinear system of ordinary differential equations (ODE) for $\bphi = \bphi(t)$:
\begin{equation}
\label{eqn:tdksODE}
i \frac{d}{dt} \bphi = \mathcal{K} \bphi + V(\bphi, \bv^C) \bphi,
\end{equation}
where $V(\bphi, \bv^C)$ is a diagonal matrix whose diagonal is the vector
$\bv(\bphi, \bv^{C}) = -((\bx + 10)^2 + 1)^{-1/2} + W  ( |\bphi|^2 \circ \bw ) + \bv^{C}$.
Here $\bx$ is the vector whose $j$-th entry is $x_j$ and operations involving $\bx$ should be interpreted entry-wise.  We have deliberately kept $\bv^{C}$ general to encompass both the cases where (i) $\bv^{C}(t)$ is a vector of time-dependent parameters whose $j$-th entry is $v^C[\phi](x_j,t)$, and (ii) $\bv^{C}$ is a function that takes as input, e.g., $\{ \bphi_k, \bphi_{k-1}, \ldots \bphi_0 \}$ and produces as output the values $v^C[\phi](x_j, t_k)$.

To solve (\ref{eqn:tdksODE}), we apply operator splitting \citep{castro2004propagators}, resulting in the fully discretized propagation equation
\begin{equation}
\label{eqn:tdksDISC2}
\bphi(t_{k+1}) = \exp(-i \mathcal{K} \Delta t/2) \exp(-i V(\bphi(t_k), \bv^{C}_k) \Delta t) \exp(-i \mathcal{K} \Delta t/2) \bphi(t_k).
\end{equation}
We choose this method for two reasons.  First, the propagator is unitary and hence preserves the normalization of $\bphi$ over long times.  Second, as $V$ is diagonal, both the matrix exponential of $-i V \Delta t$ and its Jacobian with respect to $\bv^C$ are simple to calculate. The ease with which we can compute derivatives of the right-hand side of (\ref{eqn:tdksDISC2}) balances its second-order accuracy in time.

Note that $\mathcal{K}$ is time-independent and symmetric. For small systems,  $\exp(-i \mathcal{K} \Delta t/2)$ can be computed by diagonalizing $\mathcal{K}$.  If $\mathcal{K} = S D S^{-1}$, then $\exp(-i \mathcal{K} \Delta t/2) = S \exp(-i D \Delta t/2) S^{-1}$.  During the initial part of our codes, we compute this kinetic propagator once and store it for future use.  The adjoint derivations below can be extended straightforwardly to higher-order version of operator splitting, as long as the discrete propagation scheme involves alternating products of kinetic and potential propagation terms as in (\ref{eqn:tdksDISC2}), with $V$ and $\bv^C$ exponentiated diagonally.

\subsection{Solving the Spatially Two-Dimensional (2D) Schr\"odinger System}
\label{sect:schroforward}
To generate training data, we numerically solve the 2D Schr\"odinger model system with Hamiltonian $\hat{H} = \hat{K} + \hat{V}$.  Here the kinetic operator is $\hat{K} = -(1/2) \nabla^2$, and $\nabla^2$ is the two-dimensional Laplacian.  The electronic potential $\hat{V}$ consists of a sum of electron-nuclear and electron-electron terms:
\begin{equation}
\label{eqn:schropot}
\hat{V}(x_1, x_2) = v^\text{ext}(x_1) + v^\text{ext}(x_2) + W^{ee}(x_1, x_2),
\end{equation}
with $v^\text{ext}$ and $W^{ee}$ defined via the soft-Coulomb potentials (\ref{eqn:vextdef}) and (\ref{eqn:Weedef}), respectively.  

For forward time-evolution of the 2D Schr\"odinger model system, we use a discrete Laplacian $\Delta_2$ that consists of
\[
\Delta_2 = \Delta \otimes I + I \otimes \Delta,
\]
where $I$ is the $(J+1) \times (J+1)$ identity matrix and $\otimes$ denotes the Kronecker product.  The resulting $\Delta_2$ is a fourth-order approximation to the two-dimensional Laplacian $\nabla^2 = \partial_{x_1 x_1} + \partial_{x_2 x_2}$.  In the Schr\"odinger system, spatially discretizing the kinetic operator yields the matrix $\mathcal{K} = -(1/2) \Delta_2$, which is of dimension $(J+1)^2 \times (J+1)^2$ with $J = 1200$.  To compute the kinetic portion of the propagator,
\[
\mathcal{P}_\mathcal{K} = \exp(-i \mathcal{K} \Delta t/2),
\]
we used a straightforward fourth-order series expansion of the matrix exponential:
\[
\mathcal{P}_\mathcal{K} \approx \sum_{j=0}^4 \frac{ (-i \mathcal{K} \Delta t/2)^j }{j!}.
\]
With $\Delta t = 2.4 \times 10^{-5}$ fs (or, in a.u., $\Delta t \approx 9.99219 \times 10^{-4}$), this series approximation of the matrix exponential incurs negligible error.

The potential portion of the propagator, $\mathcal{P}_V = \exp(-i V \Delta t)$, is a purely diagonal matrix---the $(J+1)^2$ entries along its diagonal consist of a flattened version of the $(J+1) \times (J+1)$ matrix obtained by evaluating (\ref{eqn:schropot}) on our finite-difference spatial grid.

Equipped with $\mathcal{P}_\mathcal{K}$ and $\mathcal{P}_V$, both of which are time-independent, we propagate forward using second-order operator splitting as in (\ref{eqn:discprop}):
\begin{equation}
\label{eqn:schroprop}
\bpsi_{k+1} = \mathcal{P}_\mathcal{K} \mathcal{P}_V \mathcal{P}_\mathcal{K} \bpsi_k.
\end{equation}
As a numerical method for the TDSE (\ref{eqn:tdse}), operator splitting goes back at least to the work of \cite{fleck1976time} and \cite{feit1982solution}.  Starting from an initial condition $\bpsi_0$ represented as a complex vector of dimension $(J+1)^2$, we iterate for $K=30000$ steps until we reach a final time of $0.72$ fs.  We have implemented the above Schr\"odinger solver using sparse linear algebra and CuPy.

\paragraph{Initializing TDKS Simulations with Memory.} When we solve (\ref{eqn:tdks1d}) with a correlation potential with memory, \emph{e.g.}, $v^C[\phi]$ that depends on both $\phi(x,t)$ and $\phi(x,t - \Delta t)$, how do we initialize the simulation?  Our solution is to start with the wave function data generated by solving the TDSE (as above).  With this data, we apply the methods from \citet[Appendix E]{ullrich2011time-dependent} to compute exact Kohn-Sham states $\bphi_k$ corresponding to $k=0$ and $k=1$ with $\Delta t = 2.4 \times 10^{-3}$ fs.  We use the exact $\bphi_0$ and $\bphi_1$ to initialize our TDKS simulations when we use a $v^C$ model with memory.

\subsection{Derivation of the Adjoint System}
\label{sect:adjderiv}
Variations of (\ref{eqn:lag}) with respect to $\bLambda^R$ and $\bLambda^I$ give the real and imaginary parts of the equality constraint (\ref{eqn:tdksDISC2}).  For the variation with respect to $\bPhi^R$, we obtain
\begin{align*}
&\delta \mathscr{L} = \frac{d}{d\epsilon} \biggr|_{\epsilon=0} \mathscr{L}(\bPhi^R + \epsilon \delta \bPhi^R, \bPhi^I, \bLambda^R, \bLambda^I, \bv^{C}) \\
 &= \sum_{k=0}^{K} \sum_{j=0}^{J}  ( 2 \phi^R(x_j, t_k)^2 + 2 \phi^I(x_j, t_k)^2 - \tilde{n}(x_j, t_k) )( 4 \phi^R(x_j, t_k) \delta \phi^R(x_j,t_k) ) \\
 &\quad - \sum_{k=0}^{K-1}  [\blambda_{k+1}^R]^T \delta \bphi_{k+1}^R  \\
 &\quad + \sum_{k=0}^{K-1} [\blambda_{k+1}^R]^T  \nabla_{\bphi^R} \bF^R_{\Delta t}(\bphi_k^R, \bphi_k^I, \bv^{C}_k) \delta \bphi^R_k  + [\blambda_{k+1}^I]^T  \nabla_{\bphi^R} \bF^I_{\Delta t}(\bphi_k^R, \bphi_k^I, \bv^{C}_k) \delta \bphi^R_k    \\
&= \sum_{k=0}^K 4 \left[ (2 |\bphi_k|^2 - \tilde{\bn}_k) \circ \bphi_k^R \right]^T \delta \bphi^R_k  - \sum_{k=1}^{K} [\blambda_{k}^R]^T \delta \bphi_{k}^R + \sum_{k=0}^{K-1} \begin{bmatrix} \blambda_{k+1}^R  \\  \blambda_{k+1}^I \end{bmatrix}^T  \nabla_{\bphi^R} \begin{bmatrix} \bF_{\Delta t}^R  \\
\bF_{\Delta t}^I \end{bmatrix}
\delta \bphi_{k}^R.
\end{align*}
Analogously, for the variation with respect to $\bPhi^I$, we obtain
\[
\delta \mathscr{L} = \sum_{k=0}^K 4 \left[ (2 |\bphi_k|^2 - \tilde{\bn}_k) \circ \bphi_k^I \right]^T \delta \bphi^I_k  - \sum_{k=1}^{K} [\blambda_{k}^I]^T \delta \bphi_{k}^I + \sum_{k=0}^{K-1} \begin{bmatrix} \blambda_{k+1}^R  \\  \blambda_{k+1}^I \end{bmatrix}^T  \nabla_{\bphi^I} \begin{bmatrix} \bF_{\Delta t}^R  \\
\bF_{\Delta t}^I \end{bmatrix}
\delta \bphi_{k}^I.
\]
Setting $\delta \mathscr{L} = 0$ for all variations $\delta \bphi_k^R$ and $\delta \bphi_k^I$ for $k \geq 1$, we obtain the following backward-in-time system for $\bLambda$:
\begin{subequations}
\label{eqn:adjsysappendix}
\begin{align}
\label{eqn:adjfinappendix}
\blambda_K &= 4 \left[ (2 |\bphi_K|^2 - \tilde{\bn}_K) \circ \bphi_K \right] \\
\label{eqn:adjReqn}
[\blambda_{k}^R]^T &= 4 \left[ (2 |\bphi_k|^2 - \tilde{\bn}_k) \circ \bphi_k^R \right]^T + \begin{bmatrix} \blambda_{k+1}^R  \\  \blambda_{k+1}^I \end{bmatrix}^T  \nabla_{\bphi^R} \begin{bmatrix} \bF_{\Delta t}^R  \\
\bF_{\Delta t}^I \end{bmatrix} (\bphi_k^R, \bphi_k^I, \bv^{C}_k)\\
\label{eqn:adjIeqn}
[\blambda_{k}^I]^T &= 4 \left[ (2 |\bphi_k|^2 - \tilde{\bn}_k) \circ \bphi_k^I \right]^T + \begin{bmatrix} \blambda_{k+1}^R  \\  \blambda_{k+1}^I \end{bmatrix}^T  \nabla_{\bphi^I} \begin{bmatrix} \bF_{\Delta t}^R  \\
\bF_{\Delta t}^I \end{bmatrix} (\bphi_k^R, \bphi_k^I, \bv^{C}_k)
\end{align}
\end{subequations}
We can write the Jacobian as a block matrix:
\begin{equation}
\label{eqn:blockjacob}
\bJ_{\bphi} \bF_{\Delta t} = \begin{bmatrix} \nabla_{\bphi^R} \bF^R_{\Delta t} & \nabla_{\bphi^I} \bF^R_{\Delta t} \\
\nabla_{\bphi^R} \bF^I_{\Delta t} & \nabla_{\bphi^I} \bF^I_{\Delta t}
\end{bmatrix}.
\end{equation}

\subsection{Gradients of the TDKS Propagator}
\label{sect:TDKSgrads}
Here we consider gradients of the propagator $\bF_{\Delta t}$ defined in (\ref{eqn:discprop}).  Note that $\bF_{\Delta t}$ also satisfies
\begin{equation*}
\bF_{\Delta t}(\bphi, \bv^{C}) = \bF^R_{\Delta t}(\bphi^R, \bphi^I, \bv^{C}) + i \bF^I_{\Delta t}(\bphi^R, \bphi^I, \bv^{C}).
\end{equation*}

\paragraph{Gradients of $\bF$ when we seek pointwise values of $\bv^C$.} Because $V$ is diagonal, the $\ell$-th element of $\bF$ is
\[
\left[ \bF_{\Delta t}(\bphi, \bv^C) \right]_{\ell} = \sum_{q, r} \mathcal{P}_{\mathcal{K};\ell,q} \exp(-i V_{qq}(\bphi,\bv^C) \Delta t) \mathcal{P}_{\mathcal{K};q, r} \phi_r.
\]
First let us compute the derivative of this $\ell$-th element with respect to $\phi_m^R$.  We obtain
\begin{multline}
\label{eqn:dFdphiR}
\frac{\partial \left[ \bF_{\Delta t}(\bphi, \bv^C) \right]_{\ell} }{\partial \phi_m^R} = \sum_{q, r} \mathcal{P}_{\mathcal{K};\ell,q} \exp(-i V_{qq}(\bphi,\bv^C) \Delta t) (-i \Delta t) \frac{\partial V_{qq}}{\partial \phi_m^R} \mathcal{P}_{\mathcal{K};q, r} \phi_r \\
+ \sum_{q} \mathcal{P}_{\mathcal{K};\ell,q} \exp(-i V_{qq}(\bphi,\bv^C) \Delta t) \mathcal{P}_{\mathcal{K};q, m},
\end{multline}
with
\[
\frac{\partial V_{qq}}{\partial \phi_m^R} = \frac{\partial}{\partial \phi_m^R} \sum_{s} W_{q,s} w_s \phi_s \phi_s^\ast = 2 W_{q,m} w_m \phi_m^R.
\]
The derivative with respect to $\phi_m^I$ is similar:
\begin{multline}
\label{eqn:dFdphiI}
\frac{\partial \left[ \bF_{\Delta t}(\bphi, \bv^C) \right]_{\ell} }{\partial \phi_m^I} = \sum_{q, r} \mathcal{P}_{\mathcal{K};\ell,q} \exp(-i V_{qq}(\bphi,\bv^C) \Delta t) (-i \Delta t) \frac{\partial V_{qq}}{\partial \phi_m^I} \mathcal{P}_{\mathcal{K};q, r} \phi_r \\
+ \sum_{q} \mathcal{P}_{\mathcal{K};\ell,q} \exp(-i V_{qq}(\bphi,\bv^C) \Delta t) \mathcal{P}_{\mathcal{K};q, m} i,
\end{multline}
with
\[
\frac{\partial V_{qq}}{\partial \phi_m^I} = \frac{\partial}{\partial \phi_m^I} \sum_{s} W_{q,s} w_s \phi_s \phi_s^\ast = 2 W_{q,m} w_m \phi_m^I.
\]
Taking the real and imaginary parts of (\ref{eqn:dFdphiR}-\ref{eqn:dFdphiI}), we obtain all necessary elements of the block Jacobian (\ref{eqn:blockjacob}).  Next we compute the derivative of the $\ell$-th element of $\bF$ with respect to the $m$-th element of $\bv^{C}$:
\begin{equation*}
\frac{\partial  \left[ \bF_{\Delta t}(\bphi, \bv^C) \right]_{\ell} }{\partial v^C_m } =  \sum_{r} \mathcal{P}_{\mathcal{K};\ell,m} \exp(-i V_{mm}(\bphi,\bv^C) \Delta t) (-i \Delta t)  \mathcal{P}_{\mathcal{K};m, r} \phi_r,
\end{equation*}
which follows from
\[
\frac{\partial V_{qq}}{\partial  v^C_m} = \delta_{qm} = \begin{cases} 1 & q = m \\ 0 & q \neq m. \end{cases}
\]

\paragraph{Gradients of $\bF$ when we model the functional dependence of $\bv^C$ on present and past states.} The $\ell$-th element of $\bF$ is now
\[
\left[ \bF_{\Delta t}(\bphi, \bphi'; \btheta) \right]_{\ell} = \sum_{q, r} \mathcal{P}_{\mathcal{K};\ell,q} \exp(-i V_{qq}(\bphi; \bphi'; \btheta) \Delta t) \mathcal{P}_{\mathcal{K};q, r} \phi_r.
\]
Using the new expression for $V$, we derive
\begin{multline}
\label{eqn:newdFdphiR}
\frac{\partial \left[ \bF_{\Delta t}(\bphi, \bphi'; \btheta) \right]_{\ell} }{\partial \phi_m^R} = \sum_{q, r} \mathcal{P}_{\mathcal{K};\ell,q} \exp(-i V_{qq}(\bphi, \bphi'; \btheta) \Delta t) (-i \Delta t) \frac{\partial V_{qq}}{\partial \phi_m^R} \mathcal{P}_{\mathcal{K};q, r} \phi_r \\
+ \sum_{q} \mathcal{P}_{\mathcal{K};\ell,q} \exp(-i V_{qq}(\bphi, \bphi'; \btheta) \Delta t) \mathcal{P}_{\mathcal{K};q, m},
\end{multline}
with
\[
\frac{\partial V_{qq}}{\partial \phi_m^R} = 2 W_{q,m} w_m \phi_m^R + \frac{\partial v^C_q}{\partial \phi_m^R}.
\]
The derivative with respect to $\phi_m^I$ is similar:
\begin{multline}
\label{eqn:newdFdphiI}
\frac{\partial \left[ \bF_{\Delta t}(\bphi, \bphi'; \btheta) \right]_{\ell} }{\partial \phi_m^I} = \sum_{q, r} \mathcal{P}_{\mathcal{K};\ell,q} \exp(-i V_{qq}(\bphi, \bphi'; \btheta) \Delta t) (-i \Delta t) \frac{\partial V_{qq}}{\partial \phi_m^I} \mathcal{P}_{\mathcal{K};q, r} \phi_r \\
+ \sum_{q} \mathcal{P}_{\mathcal{K};\ell,q} \exp(-i V_{qq}(\bphi, \bphi'; \btheta) \Delta t) \mathcal{P}_{\mathcal{K};q, m} i,
\end{multline}
with
\[
\frac{\partial V_{qq}}{\partial \phi_m^I} = 2 W_{q,m} w_m \phi_m^I + \frac{\partial v^C_q}{\partial \phi_m^I}.
\]
The derivatives with respect to the past state $\bphi'$ are
\begin{subequations}
\label{eqn:newdFdphiprime}
\begin{align}
\frac{\partial \left[ \bF_{\Delta t}(\bphi, \bphi'; \btheta) \right]_{\ell} }{\partial \phi_m^{',R}} &=  \sum_{q, r} \mathcal{P}_{\mathcal{K};\ell,q} \exp(-i V_{qq}(\bphi, \bphi'; \btheta) \Delta t) (-i \Delta t) \frac{\partial v^C_{q}}{\partial \phi_m^{',R}} \mathcal{P}_{\mathcal{K};q, r} \phi_r \\
\frac{\partial \left[ \bF_{\Delta t}(\bphi, \bphi'; \btheta) \right]_{\ell} }{\partial \phi_m^{',I}} &=  \sum_{q, r} \mathcal{P}_{\mathcal{K};\ell,q} \exp(-i V_{qq}(\bphi, \bphi'; \btheta) \Delta t) (-i \Delta t) \frac{\partial v^C_{q}}{\partial \phi_m^{',I}} \mathcal{P}_{\mathcal{K};q, r} \phi_r
\end{align}
\end{subequations}

Taking the real and imaginary parts of (\ref{eqn:newdFdphiR}-\ref{eqn:newdFdphiI}-\ref{eqn:newdFdphiprime}), we obtain all necessary elements of both block Jacobians in (\ref{eqn:newadjiter}).

Finally, we need
\begin{equation*}
\frac{\partial \left[ \bF_{\Delta t}(\bphi, \bphi'; \btheta) \right]_{\ell} }{\partial \theta_m} = \sum_{q,r} \mathcal{P}_{\mathcal{K};\ell,q} \exp(-i V_{qq}(\bphi, \bphi'; \btheta) \Delta t) (-i \Delta t) \frac{\partial V_{qq}}{\partial \theta_m} \mathcal{P}_{\mathcal{K};q,r}\phi_r. 
\end{equation*}

\subsection{Further Implementation Details}
We implemented the adjoint method in JAX.  Derivatives of the $v^C$ model are computed via automatic differentiation.  XLA compilation enables us to run the code on GPUs.  To optimize via L-BFGS-B, we use scipy.optimize.  All source code is available upon request.

\subsection{Training Set Results}
\label{sect:trainresults}
Here we consider training a model $v^C[\phi](x,t)$ that allows for arbitrary dependence on the real and imaginary parts of both $\phi(x,t)$ and $\phi(x,t-\Delta t)$.   Here the input layer is of dimension $4(J+1)$---see Section \ref{sect:deets}.

To train such a model, we apply the L-BFGS-B optimizer with objective function given by the first line of (\ref{eqn:newlag}) and gradients computed with the adjoint system (\ref{eqn:newadj}-\ref{eqn:nablatheta}).  We initialize neural network parameters $\btheta$ by sampling a mean-zero normal distribution with standard deviation $\sigma = 0.01$.  For training data, we subsample the $p=-1.5$ TDSE electron density time series by a factor of $100$ in time, so that $\Delta t = 2.4 \times 10^{-3}$ fs and the entire training trajectory consists of $K=301$ time steps.  We retain this time step in all training sets and results that follow.

\begin{figure}
\includegraphics[width=6in]{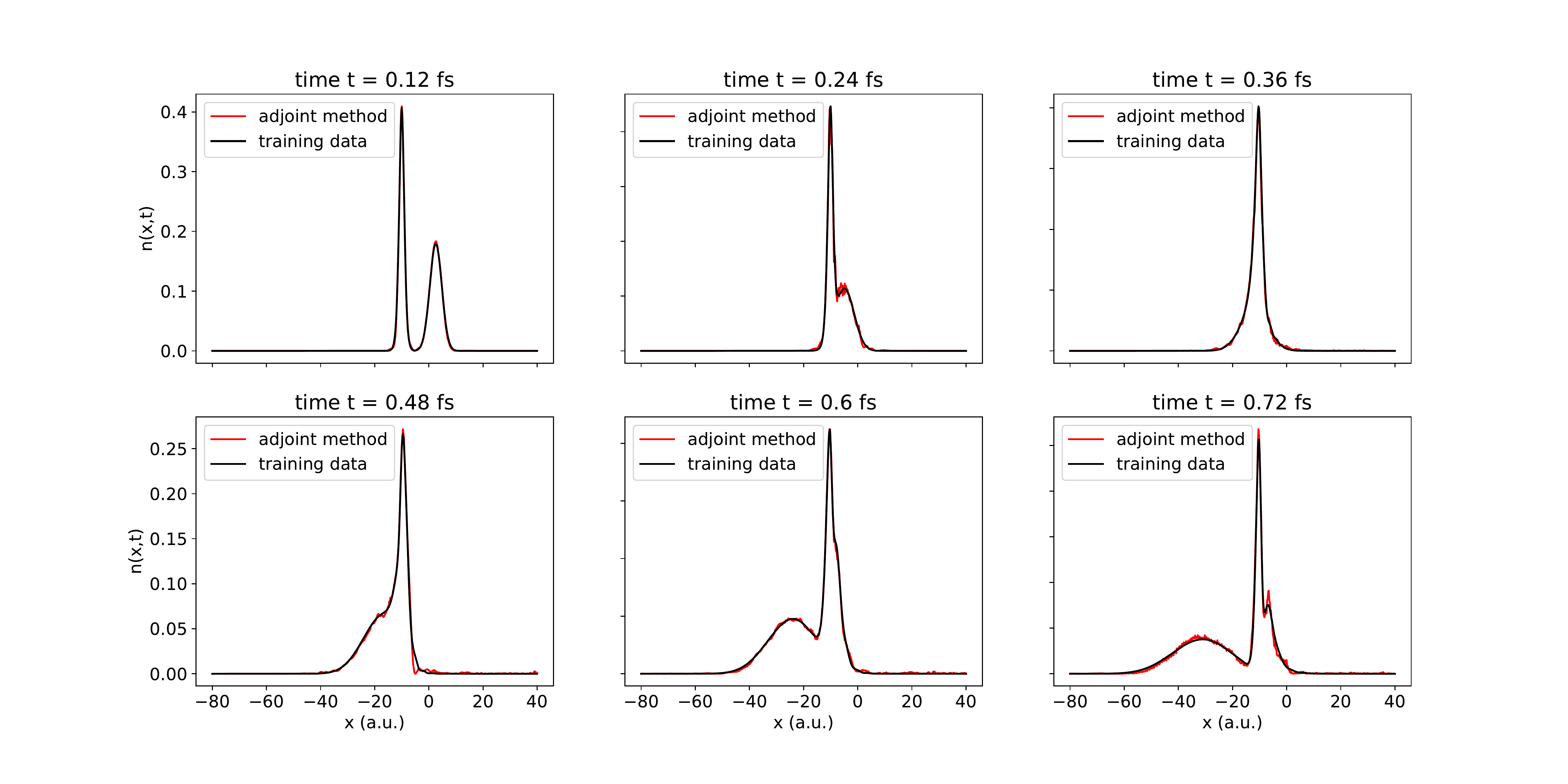}
\caption{We use $300$ time steps (corresponding to $0.72$ fs) of the $p=-1.5$ data (black) together with the adjoint method to train a neural network model $v^C[\phi]$ that depends on the current and previous $\phi$.  We plot in red the results of using the learned $v^C$ to propagate (\ref{eqn:tdks1d}) from $t=0$ to $t=0.72$ fs.  Note the close agreement.}
\label{fig:vCp15TRAIN}
\end{figure}

In Figure \ref{fig:vCp15TRAIN}, we show the resulting model's results on the training set.  The trained $v^C[\phi]$ functional, when used to solve the TDKS equation (\ref{eqn:tdks1d}), yields electron densities that agree closely with the reference TDSE electron densities.

\subsection{Relationship to Existing Literature on Optimal Control for TDKS Systems}
\label{sect:relationship}
Here we contrast our work with prior work on optimal control for TDKS systems, specifically work that involves the adjoint method.

First let us view our work through the lens of optimal control: we generate reference data by first solving the two-dimensional TDSE.  Our cost function is then the mismatch between (i) electron densities computed from TDKS, and (ii) electron densities computed from the time-dependent wave functions obtained from TDSE, all on a discrete temporal grid.  We view $v^C$ as a control that, properly chosen, guides TDKS to produce the same electron densities that would have been produced by solving TDSE.  

A common feature of both present and prior work is the idea of incorporating the TDKS equation as a time-dependent constraint---see Eq. (8) in \cite{PhysRevLett.109.153603},  Eq. (43) in \cite{castro2013optimal}, and Eq. (3.4) in \cite{sprengel2018investigation}.  Upon taking functional derivatives, this leads naturally to adjoint systems, which have been analyzed in detail for TDKS systems \citep{Borzi2012, sprengel2017theoretical}.  In particular,  \cite{sprengel2017theoretical} and \cite{sprengel2018investigation} develop and analyze optimal control problems for multidimensional TDKS systems.  In all prior work we have seen, the correlation potential $v^C$ is taken as \emph{adiabatic}  with fixed functional form throughout the solution of the optimal control problem.

In prior work, the control $u$ is distinct from $v^C$.  In \cite{PhysRevLett.109.153603}, the control $u$ governs the Fourier spectrum of the amplitudes of an applied electric field. In \cite{sprengel2018investigation}, the control $u$ influences the system through potentials such as $V_u(x) = x^2$ and $V_u(x) = x \cdot p$, modeling the control of a quantum dot.

In \cite{PhysRevLett.109.153603}, the objective is to balance (i) maximization of charge transfer from one potential well to a neighboring potential well with (ii) minimization of the intensity of the applied field.  The resulting cost function models both parts of this physical objective.  In \cite{sprengel2018investigation}, the authors do include in their cost function the $L^2$ distance between the electron density computed from TDKS and a reference electron density, all in continuous time.  They apply this to the problem of guiding TDKS towards a target trajectory that itself was computed by solving TDKS.

Viewed in this context, the distinguishing features of the present work are as follows: (i) treating $v^C$ itself as the control, (ii) allowing $v^C$ to be \emph{non-adiabatic} in the sense that it depends on both present and past electron densities, and (iii) applying this method to match electron density trajectories computed from TDSE.  By treating $v^C$ as the object of interest, and by guiding TDKS trajectories to match TDSE trajectories, the present work addresses the system identification problem of learning $v^C$ from data.

 \newpage

 \pagenumbering{arabic}
 \setcounter{page}{12}
 
\bibliography{adjtdks}
\end{document}